\pdfoutput=1

\documentclass[11pt]{article}

\usepackage[review]{acl}
\usepackage{graphicx}
\usepackage{times}
\usepackage{latexsym}
\usepackage{float}

\usepackage[T1]{fontenc}

\usepackage[utf8]{inputenc}

\usepackage{microtype}

\title{Harnessing the Power of BERT in the Turkish Clinical Domain: Pretraining Approaches for Limited Data Scenarios}

\author{Hazal Türkmen, Oğuz Dikenelli  \\
  Computer Engineering, Faculty of Engineering, Ege University, İzmir, Turkey \\
  \texttt{\{hazal.turkmen, oguz.dikenelli\}@ege.edu.tr}
  \AND
  Cenk Eraslan, Mehmet Cem Çallı, Süha Süreyya Özbek\\
  Radiology, Faculty of Medicine, Ege University, İzmir, Turkey\\
  \texttt{\{cenk.eraslan, cem.calli, sureyya.ozbek\}@ege.edu.tr}
}
\begin{document}

\nolinenumbers
{\makeatletter\acl@finalcopytrue
  \maketitle
}

\begin{abstract}
In recent years, major advancements in natural language processing (NLP) have been driven by the emergence of large language models (LLMs), which have significantly revolutionized research and development within the field. Building upon this progress, our study delves into the effects of various pre-training methodologies on Turkish clinical language models' performance in a multi-label classification task involving radiology reports, with a focus on addressing the challenges posed by limited language resources. Additionally, we evaluated the simultaneous pretraining approach by utilizing limited clinical task data for the first time. We developed four models, including TurkRadBERT-task v1, TurkRadBERT-task v2, TurkRadBERT-sim v1, and TurkRadBERT-sim v2. Our findings indicate that the general Turkish BERT model (BERTurk) and TurkRadBERT-task v1, both of which utilize knowledge from a substantial general-domain corpus, demonstrate the best overall performance. Although the task-adaptive pre-training approach has the potential to capture domain-specific patterns, it is constrained by the limited task-specific corpus and may be susceptible to overfitting. Furthermore, our results underscore the significance of domain-specific vocabulary during pre-training for enhancing model performance. Ultimately, we observe that the combination of general-domain knowledge and task-specific fine-tuning is essential for achieving optimal performance across a range of categories. This study offers valuable insights for developing effective Turkish clinical language models and can guide future research on pre-training techniques for other low-resource languages within the clinical domain.
\end{abstract}

\section{Introduction}
Language models have undergone a significant transformation in the field of natural language processing, with exceptional capabilities demonstrated in executing tasks with minimal guidance. This shift can be attributed to pivotal milestones such as word2vec \citep{mikolov2013distributed}, which replaced feature engineering methods with deep learning-based representation learning. Furthermore, the emergence of pre-trained transformer-based models such as BERT \citep{devlin2018bert}, GPT \citep{radford2018improving}, T5 \citep{raffel2020exploring}, and BART \citep{lewis2019bart} has led to the development of contextualized word embeddings with ELMo \citep{peters1802deep}\\
Recent advancements in large language models (LLMs) have led to the development of models with parameter sizes exceeding a hundred billion, such as GPT-3 \citep{rae2021scaling}, which are pre-trained on massive datasets. However, there is a scarcity of research focusing on LLMs architectures within specialized domains characterized by limited resources. A range of approaches for developing language models exists to address the issue of limited language resources, including simultaneous pretraining with in-domain data \citep{wada2020pre} and domain-adaptive pretraining by fine-tuning an existing generic language model with in-domain data \citep{gururangan2020don}. The choice of pre-training technique depends on the specific task data and available resources, but determining the optimal utilization of limited clinical task data in pretraining and selecting the most suitable data for pretraining methods remain open questions. This study aims to assess and contrast different techniques using a limited task corpus for pretraining BERT models in the Turkish clinical domain, which is low-resource settings. This work introduces four pretrained language models for the clinical domain in the Turkish language. These models explore the effects of different corpus selections combining small task-related corpus and pretraining strategies  in the Turkish clinical domain. We also created a labeled dataset for multi-label classification using head CT radiology reports to evaluate the models. The main contributions can be listed as:\\
\begin{itemize}
\item While simultaneous pretraining has previously been explored with limited biomedical literature data in the work of \citep{wada2020pre}, our study shifts the focus towards applying this approach to limited clinical Turkish radiology data for the first time. We conducted an evaluation of simultaneous pretraining, incorporating limited clinical task radiology data, and compared it with task-adaptive pretraining through continual pre-training. This novel comparison provides valuable insights into the efficacy of these methods in the context of limited clinical radiology data, highlighting their potential in specialized domains.
\item We created small task-related corpora, including Turkish head CT radiology reports by Ege University Hospital. Then, we built four  pretrained clinical language models, for the first time, using Turkish head CT radiology reports, Turkish general corpus and Turkish biomedical corpara including Turkish medical articles \citep{turkmen2022developing}, Turkish radiology theses \citep{turkmen2022developing}.
\item We developed a multi-label classification task aimed at identifying the presence or absence of 12 clinically significant observations, as well as a "no findings" label indicating no observations, within head CT radiology reports for the purpose of evaluating language models. To the best of our knowledge, there are no existing multi-label text classification studies in the Turkish clinical domain.

\end{itemize} 

\begin{table*}[htb]
\centering
\begin{tabular}{llll}
\hline
\textbf{Corpus} & \textbf{Size (GB)} & \textbf{N tokens} & \textbf{Domain}\\
\hline
General Turkish Corpus & 35 & 4,404,976,662 & General \\
Turkish Biomedical Corpus & 0,48 & 60,318,554 & Biomedical\\
Turkish Electronic Radiology Theses & 0,11 &15,268,779 & Radiology \\
Head CT Reports & 0.036 & 4,177,140 & Clinical Radiology \\
\hline
\end{tabular}
\caption{\label{tab:corpora}
Corpora statistics}
\end{table*}

\section{Related Work}
In the pursuit of optimizing natural language processing models for specialized domains, various studies have explored different approaches to adapt general BERT models for the biomedical domain. BioBERT \citep{lee2020biobert}, an early attempt to adapt general BERT models to the biomedical domain, employed continual pretraining to enhance performance. Initialized from the general BERT model, BioBERT was further trained on PubMed abstracts and full-text articles, yielding improved performance for tasks like named entity recognition, relation extraction, and question answering. Similarly, ClinicalBERT \citep{alsentzer2019publicly}, a domain-specific language model, was created using continual pretraining with MIMIC data, demonstrating its effectiveness in improving clinical task performance.\\
Other studies explored continual pretraining for biomedical language models, such as SciBERT \citep{beltagy2019scibert} and BlueBERT \citep{beltagy2019scibert}, which were pretrained on a mix of biomedical and general domain corpora. An alternative approach, pretraining from scratch, focuses on in-domain data exclusively without relying on a generic language model. This method has been effective in creating models like PubMedBERT \citep{gu2021domain}, which is pretrained solely on PubMed abstracts. Comparisons between the two pretraining methods reveal that continual pretraining often leads to more successful transfers from general to specialized domains. For example, a study proposed four BERT models \citep{bressem2020highly}, two pretrained on German radiology free-text reports (FS-BERT, RAD-BERT) and two based on open-source models (MULTI-BERT, GER-BERT). The FS-BERT model, which used the pretraining from scratch approach, performed poorly compared to other models, suggesting that domain-specific corpora alone might be insufficient for learning proper embeddings. Another study developed RadBERT \citep{yan2022radbert}, a set of six transformer-based language models pretrained on radiology reports with various language models as initialization, exploring their performance in radiology NLP applications.\\
Although pretraining BERT models can improve performance across various biomedical NLP tasks, it requires significant domain-specific data. Biomedical text data is often limited and scattered across various sources, and few publicly available medical databases are written in languages other than English. This creates a high demand for effective techniques that can work well even with limited resources.  One solution to this problem is the Simultaneous pre-training technique proposed in the study \citep{wada2020pre}, to up-sample a limited domain-specific corpus and use it for pre-training in a balanced manner with a larger corpus.  Using small Japanese medical article abstracts and  Japanese Wikipedia text, the authors created a simultaneous pre-trained BERT model, ouBioBERT. The study confirmed that their Japanese medical BERT model performed better than the conventional baselines and other BERT models in a medical Japanese document classification task. However, they did not focus on applying the simultaneous pre-training approach to limited clinical task radiology data.Building upon this work, our study shifts the focus towards applying the simultaneous pre-training approach to limited clinical task data for the first time. Another solution to overcome the barriers of the limited resource problem, many researchers explore the benefits of continued pre-training on a smaller corpus drawn from the task distribution as task-adaptive pre-training \citep{gururangan2020don,schneider2020biobertpt}. Also \citep{turkmen2022bioberturk} previously demonstrated that their biomedical BERT models, the BioBERTurk family, which were continuously pre-trained on a limited Turkish radiology thesis corpus, exhibited improved performance in clinical tasks. However, the authors also highlighted the potential ineffectiveness of domain incompatibility when evaluating Turkish language models, emphasizing the need for a closer alignment between the domain-specific data and the evaluation tasks.

\section{Materials and Methods}
In this section, we provide a concise overview of the pre-training methods employed for the development of Turkish clinical language models and the characteristics of the corpora utilized in this process. We developed four Turkish clinical language models, leveraging the BERT-base architecture and constrained language resources, by employing two pre-training strategies: simultaneous pre-training and contiunal pre-training, referred to as task-adaptive pretraining. Two models, referred to as the TurkRadBERT-sim family, were developed employing simultaneous pre-training techniques that combined general, biomedical, and clinical task corpora while utilizing distinct vocabularies. In contrast, two models, the TurkRadBERT-task family, were developed employing task-adaptive pre-training using task corpus. To construct these clinical models, we employed four distinct corpora: the Turkish biomedical corpus compiled from open-source medical articles \citep{turkmen2022developing}, the Turkish electronic radiology theses corpus \citep{turkmen2022developing}, the Turkish web corpus \citep{stefan-it}, and the newly created Turkish radiology reports corpus which is limited task corpus. While all corpora were utilized in simultaneous pre-training, only the Turkish radiology reports were used in task-adaptive pre-training. Subsequently, the clinical language models were fine-tuned on a downstream NLP task within the Turkish clinical domain. Finally, the clinical language models were compared to the general Turkish domain BERT model, BERTurk \citep{stefan-it}.

\subsection{Pre-training Strategies}
 The BERT framework \citep{devlin2018bert} consists of two phases: pre-training and fine-tuning. During pre-training, BERT is trained on large-scale plain text corpora such as Wikipedia, whereas in the fine-tuning phase, it is initialized with the same pre-trained weights and then fine-tuned using task-specific labeled data, such as sentence pair classification. BERT employs two unsupervised tasks during the pre-training phase: Masked Language Model (MLM) and Next Sentence Prediction (NSP). In MLM tasks, a certain percentage of input tokens are randomly masked, and the model predicts the masked tokens in a sentence, as described in the Cloze task \citep{taylor1953cloze}. For NSP, the model predicts whether the second sentence follows a consecutive sentence in the dataset.\\
In our study, we implemented several modifications to the BERT architecture for simultaneous pre-training \citep{wada2020pre}, our first technique. This pre-training approach posits that training the BERT model using large and small corpora together can prevent overfitting issues caused by limited medical data. To accurately feed inputs to the model, we followed a procedure same study \citep{wada2020pre}. We divided the small medical corpus and large general corpus into smaller documents of equal size and combined them to create structured inputs. This approach mitigates potential overfitting resulting from limited data size by increasing the frequency of pre-training for MLM instances containing small medical data. In accordance with the same study by \citep{wada2020pre}, we utilized domain-specific generated text and the Wordpiece algorithm to generate a domain-specific vocabulary, which is referred to as an amplified vocabulary in their research. Thus, we examined the impact of the domain-specific vocabulary.\\
Simultaneous pre-training enables the model to learn language representations by training on large-scale text. However, this approach can be expensive due to the extensive amount of data involved. Lastly, we implemented the task-adaptive pre-training method \citep{gururangan2020don} using only small clinical task data. This technique is less resource-intensive compared to the others. In contrast to the aforementioned pre-training methods, we developed different BERT models based on model initialization for task-adaptive pre-training, opting to use the existing BERT vocabulary instead of creating a new one.

\subsection{Data Sources for Model Development}
In the development of various language models, multiple corpora were utilized to ensure that the models were well suited to the specific domain and task at hand. The selection of appropriate corpora is crucial to the performance of language models, as it directly influences their understanding of domain-specific language patterns, structures, and vocabularies.
The corpora used are summarized in Table \ref{tab:corpora} and listed below:\\
\textbf{Head CT Reports:} We collected 40,306 verified Turkish radiology reports pertaining to computed tomography (CT) examinations for patients aged 8 years and above from the neurology and emergency departments at Ege University Hospital between January 2016 and June 2018. Prior to data analysis, reports containing fewer than 100 characters were excluded, and newline characters and radiology-specific encodings were removed for consistency. All text data underwent de-identification and duplicate removal. Following preprocessing, 2,000 reports were randomly selected for the head CT annotation task, and the remaining data (approximately 36 MB) was reserved for pre-training techniques. \\
\textbf{General Turkish Corpus:} This corpus, which was used in the development of the BERTurk model, contains a large collection of Turkish text data (approximately 35 GB). This serves as a foundation for training language models to understand Turkish language patterns.\\
\textbf{Turkish Biomedical Corpus:} A domain-specific corpus \citep{turkmen2022developing} consisting of full-text articles collected from Dergipark, a platform hosting periodically refereed biomedical journals in Turkey.\\
\textbf{Turkish Electronic Radiology Theses:} A unique corpus of open-domain Ph.D. theses \citep{turkmen2022developing} conducted in radiology departments of medical schools obtained from the Turkish Council of Higher Education's website.

\subsection{Data preparation}
The first phase after data understanding is transforming the text to the BERT-supported inputs, namely tokenization. All engineering processes to be fed into BERT were designed for Google Cloud TPUs and implemented using CPU core i8. Furthermore, Wordpiece algorithm was used to generate vocabulary for tokenization in both pre-training methods due to the success in morphologic-rich languages such as Turkish \citep{toraman2023impact}. Each vocabulary config file is the same as BERTurk for a fair comparison. We implemented the tokenizer library from Huggignface \footnote{https://huggingface.co/docs/tokenizers/python/latest/}  to build BERT’s vocabulary in simultaneous pre-training and pre-training from scratch. For continual pre-training, we used existing BERT’s vocabulary for continual pre-training instead of creating a new one. After this process, we used \textit{create\_pretraining\_data.py} script provided by the Google AI Research team \footnote{https://github.com/google-research/bert} to convert all documents into TensorFlow examples compatible with TPU devices.

\subsection{Pretraining setup}
We followed BERT-base architecture consisting of 12 layers of transformer blocks, 12 attention heads, and 110 million parameters for all pre-training strategies. All models were also generated using the same hyperparameters (see Appendix~\ref{sec:appendixB}, Table \ref{tab:pretraining} ) and were trained  with open-source training scripts available in the official BERT GitHub repository using V3 TPUs with 32 cores from Google Cloud Compute Services \footnote{https://cloud.google.com/}.

\subsection{Developed Language Models}

The simultaneous pre-training technique is the first pre-training method we implemented to utilize a small in-domain corpus.
Moreover, the first step in simultaneous pre-training is choosing data for small and large corpus data. We produced different TurkRadBERT-sim models according to the corpus selection. \\
\textbf{TurkRadBERT-sim v1} employed a large Turkish general corpus (35 GB) used for developing BERTurk, alongside a mixed Turkish biomedical corpus, Turkish Electronic Radiology Theses, and Turkish Head CT Reports as smaller counterparts. Excluding the data utilized for labeling (approximately 6 MB), the head CT reports were not used as a standalone small corpus for pre-training due to their limited size (30 MB) compared to other corpora. Furthermore, experimental results suggested that simultaneous training with such a data size did not yield significant outcomes in radiology report classification. To address this, we combined the small-sized corpus to match the large one, creating pre-training instances. The model also employed an amplified vocabulary, built from the generated corpus, for simultaneous pre-training.\\
\textbf{TurkRadBERT-sim v2} was also based on the BERT-base architecture and was pre-trained simultaneously. The model used the same corpus as v1 during pretraining. The difference was that the general domain vocabulary was used to observe the effect of the domain-specific vocabulary.\\
The last pre-training method is task-adaptive pre-training on radiology reports (30 MB). We developed two different BERT models according to the model initialization.\\
\textbf{TurkRadBERT-task v1} used a general domain language model for Turkish, BERTurk for model initialization and then carried out continual pre-training as a task-adaptive pre-training method. Vocabulary was also inherited from BERTurk.\\
\textbf{TurkRadBERT-task v2} used a Turkish biomedical BERT model, BioBERTurk variant\citep{turkmen2022bioberturk}, which was further pre-trained on Turkish electronic theses for model initialization. This Turkish biomedical BERT was chosen because it achieved the best score in classification radiology reports \citep{turkmen2022bioberturk}. For tokenization, the model again inherited from the general domain.\\

\begin{table*} [ht!]
\centering
\begin{tabular}{lccc}
\hline
\textbf{Model} & \textbf{Precision} & \textbf{Recall} & \textbf{F1 Score} \\
\hline
BERTurk              & 0.9738    & 0.9456 & 0.9562   \\
TurkRadBERT-task v1  & 0.9736    & 0.9462 & 0.9556   \\
TurkRadBERT-task v2  & 0.9643    & 0.9352 & 0.9470   \\
TurkRadBERT-sim v1   & 0.8613    & 0.7969 & 0.8149   \\
TurkRadBERT-sim v2   & 0.8170    & 0.7863 & 0.7879   \\ 
\hline
\end{tabular}
\caption{Precision, recall, and F1 Score for each model}
\label{table:model_metrics}
\end{table*}

\begin{table*} [ht!]
\centering
\begin{tabular}{lcc}
\hline
\textbf{Category} & \textbf{BERTurk} & \textbf{TurkRadBERT-task v1} \\
\hline
Intraventricular & 0.4815 & 0.4000 \\
Gliosis & 0.8580 & 0.8155 \\
Epidural & 0.9012 & 0.9000 \\
Hydrocephalus & 0.9458 & 0.9673 \\
Encephalomalacia &  0.9622 & 0.9633 \\
Chronic ischemic changes & 0.9918 & 0.9921 \\
Lacuna & 0.9655 & 0.9655 \\
Leukoaraiosis & 0.8995 & 0.8762 \\
Mega cisterna magna & 0.6000 & 0.4500 \\
Meningioma & 1.0000 & 1.0000 \\
Subarachnoid Bleeding & 0.9281 & 0.9544 \\
Subdural & 0.9666 & 0.9757 \\
No Findings & 0.9455 & 0.9311 \\

\hline
\end{tabular}
\caption{F1 scores for each label in the TurkRadBERT-task v1 and BERTurk models}
\label{table:category_f1_scores}
\end{table*}

\section{Supervision Task}

\subsection{Multi-label CT radiology reports classification}
We developed a multi-label document classification task using 2000 Turkish head CT reports mentioned in section 3.2. This was necessary as there was no shared task for clinical documents in Turkish. Our dataset has 20618 sentences and 249072 tokens. The objective of the reports classification task is to identify the existence of clinically significant observations in a radiology report that is presented in free-text format. These are 'Intraventricular' ,'Gliosis', 'Epidural', 'Hydrocephalus', 'Encephalomalacia', 'Chronic ischemic changes', 'Lacuna', 'Leukoaraiosis', 'Mega cisterna magna' ,'Meningioma', 'Subarachnoid Bleeding', 'Subdural', 'No Findings'. The classification process involves reviewing sentences within the report and categorizing them into one of two classes: positive or negative. The 13th observation, “No Findings”, indicates the absence of any findings. Radiology experts labeled the dataset according to this annotation schema. The annotation process unfolded in three stages, involving three experienced radiologists (C.E, M.C.C, and S.S.O). In each stage, two annotators (C.E, M.C.C) independently labeled a portion of the reports. Subsequently, the third annotator examined these annotations to detect any discrepancies. At the conclusion of each stage, all three annotators reached a consensus by generating mutually agreed-upon annotations. A spreadsheet file was utilized to facilitate the annotation task for the annotators The annotated datasets were subsequently divided randomly into test (10\%), validation (10\%), and training (80\%) sets for fine-tuning. The class distributions, as illustrated in Appendix~\ref{sec:appendixA}, demonstrate the varying prevalence of different categories in the datasets. The datasets exhibit an imbalanced distribution, which is a typical characteristic of text processing in the radiology domain \citep{qu2020assessing}.
\subsection{Fine-tuning Setup}
The fine-tuning of all pretrained models was conducted independently utilizing identical architecture and optimization methods as previously employed in the study \citep{devlin2018bert}. In the process of fine-tuning, the objective is not to surpass the current state-of-the-art performance on the downstream tasks, but rather to assess and compare pretraining techniques for developing  Turkish clinical language models. So, an exhaustive exploration of hyperparameters was not utilized. The configurations employed for the TurkRadBERT-sim and TurkRadBERT-task models are displayed in Table \ref{tab:finetuningsim} and Table \ref{tab:finetuningtask} in Appendix ~\ref{sec:appendixB} respectively.\\
To assess the performance of the different pre-trained Turkish clinical BERT models on the clinical multilabel classification task, we calculated the average precision, average recall, and average F1 score for each model.

\section{Experimental Results}
In this study, we evaluated the performance of five different models, including BERTurk, TurkRadBERT-task v1, TurkRadBERT-task v2, TurkRadBERT-sim v1, and TurkRadBERT-sim v2, for Turkish clinical multi-label classification. We compared their performance over ten runs in terms of average precision, recall, and F1 score. Additionally, we analyzed the performance  of wining two model (BERTurk, TurkRadBERT-task v1) on individual categories using their respective F1 scores. The results are presented in Tables \ref{table:model_metrics} and \ref{table:category_f1_scores}.\\
Table \ref{table:model_metrics} shows that BERTurk achieves an F1 score of 0.9562, with a precision of 0.9738 and recall of 0.9456. TurkRadBERT-task v1 has a slightly lower F1 score of 0.9556 but with comparable precision (0.9736) and recall (0.9462). Both models demonstrate strong performance on the classification task, with BERTurk slightly outperforming TurkRadBERT-task v1 in terms of the overall F1 score. While BERTurk performed  better than TurkRadBERT-task v1, there are no statistical differences between these models (P value 0,255). Other models, such as TurkRadBERT-task v2, TurkRadBERT-sim v1, and TurkRadBERT-sim v2, show lower overall performance compared to BERTurk and TurkRadBERT-task v1.\\
However, it is essential to evaluate the models' performance for each label, as this offers a deeper understanding of their strengths and weaknesses. Table \ref{table:category_f1_scores} presents the F1 scores for each category for BERTurk and TurkRadBERT-task v1. The results reveal that the performance of the models varies across categories, with some labels showing a noticeable difference in F1 scores between the two models. BERTurk performs better than TurkRadBERT-task v1 in the following categories: Intraventricular, Gliosis, Epidural, Leukoaraiosis, Mega cisterna magna, and No Findings. In contrast, TurkRadBERT-task v1 outperforms BERTurk in the categories of Hydrocephalus, Encephalomalacia, Chronic ischemic changes, Subarachnoid Bleeding, and Subdural. The F1 scores for Lacuna and Meningioma are identical for both models.

\section{Discussion}
Upon assessing the experiments as a whole, we derive the following conclusions. When comparing simultaneous pre-training and task-adaptive pre-training, it is observed that due to the size difference between the task data and the general data, the limited domain-specific data may be overshadowed by the large general-domain data, causing the model to focus more on learning general features rather than task-specific features. This phenomenon highlights the importance of carefully balancing the general and domain-specific data during the pre-training process to ensure that the model effectively captures the nuances of the specialized domain.\\
The performance of BERTurk and TurkRadBERT-task v1 models is quite close because both models leverage the knowledge gained from the large general-domain corpus during pre-training. BERTurk is directly pre-trained on this large corpus, while TurkRadBERT-task v1 is initialized with BERTurk's weights and then fine-tuned using task-adaptive pre-training on a smaller clinical corpus. This fine-tuning enables TurkRadBERT-task v1 to capture domain-specific patterns, structures, and terminologies absent in the general-domain corpus.\\
However, the small task-specific corpus used in task-adaptive pre-training may limit the model's learning of domain-specific knowledge. Consequently, despite the benefits of task-adaptive pre-training, TurkRadBERT-task v1 (which utilized this approach) has slightly lower performance than BERTurk. In limited data scenarios, the task-adaptive pre-training approach might be prone to overfitting, especially when pre-trained on a small task-specific corpus. The model may become overly specialized to the training data and fail to generalize well on unseen examples \citep{zhang2022shifting}.\\
In terms of performance, TurkRadBERT-task v1 has a slightly higher F1 score (0.9556) than TurkRadBERT-task v2 (0.9470). This suggests that, despite the more specialized biomedical knowledge in BioBERTurk, the general-domain BERTurk model provides a more robust foundation for task-adaptive pre-training in this specific clinical task.\\
Another conclusion reached in this study is that comparison between TurkRadBERT-sim v1 and v2 offers insights into the impact of domain-specific vocabulary on model performance. TurkRadBERT-sim v1, which used an amplified vocabulary built from the generated corpus, outperformed TurkRadBERT-sim v2 that employed the general domain vocabulary. This finding indicates that using a domain-specific vocabulary during pre-training can enhance the ability of the model to capture and understand domain-specific language patterns, ultimately leading to improved performance on clinical NLP tasks.\\
Examining the F1 scores for each label in Table \ref{table:category_f1_scores} provides a more detailed perspective on for the two most successful models performance. BERTurk outperforms TurkRadBERT-task v1 in certain labels, such as Intraventricular, Gliosis, Epidural, Leukoaraiosis, Mega cisterna magna, and No Findings. The higher performance of BERTurk on certain labels could be attributed to the general-domain knowledge it acquires during pre-training, which may provide better coverage for specific categories, particularly those with lower frequency in the task-specific corpus. BERTurk's broader pre-training data exposure could potentially give it an advantage over models like TurkRadBERT-task v1 when dealing with specific labels that have lower representation in the task-specific corpus, even though TurkRadBERT-task v1 is initialized with BERTurk. This suggests that the combination of general-domain knowledge and task-specific fine-tuning may be critical for optimal performance across diverse categories. On the other hand, TurkRadBERT-task v1 exhibits superior performance for labels like Hydrocephalus, Encephalomalacia, Subarachnoid Bleeding, and Subdural. This suggests that task-adaptive pre-training can offer a performance boost in some instances by fine-tuning the model on domain-specific information. However, it is worth noting that the overall performance differences between the two models are relatively small, highlighting the importance of leveraging both general-domain and task-specific knowledge in these models.

\section{Conclusion}
In conclusion, this study provides a comprehensive comparison of the performance of various models, including BERTurk, TurkRadBERT-task v1, TurkRadBERT-task v2, TurkRadBERT-sim v1, and TurkRadBERT-sim v2, on a radiology report classification task. Our findings demonstrate that the BERTurk model achieves the best overall performance, closely followed by the TurkRadBERT-task v1 model. This highlights the importance of leveraging both general-domain knowledge acquired during pre-training and task-specific knowledge through fine-tuning to achieve optimal performance on complex tasks.

We also observe that the performance of these models varies across different labels, with BERTurk performing better on certain categories, particularly those with lower representation in the task-specific corpus. This suggests that a combination of general-domain knowledge and task-specific fine-tuning may be critical for achieving optimal performance across diverse categories. Additionally, it is essential to consider label frequencies when interpreting results, as performance on rare labels may be more susceptible to noise and overfitting.\\
The simultaneous pre-training models, TurkRadBERT-sim v1 and v2, exhibit lower performance compared to their task-adaptive counterparts, indicating that task-adaptive pre-training is more effective in capturing domain-specific knowledge. Nevertheless, further investigation into alternative pre-training and fine-tuning strategies could help enhance the performance of these models.\\
Future research could focus on expanding the task-specific corpus to improve domain-specific knowledge and performance on rare labels, as well as exploring alternative pre-training and fine-tuning strategies to further enhance model performance. Moreover, investigating the factors contributing to the performance differences between models for each label could provide valuable insights for developing more effective models in the field of medical natural language processing.

\section*{Acknowledgements}
The study was approved by the Ege University Ethical Committee under study number UH150040389 and conducted in accordance with the Declaration of Helsinki. We would also like to express our gratitude to the TPU Research Cloud program (TRC) \footnote{https://sites.research.google/trc/about/} and Google's CURe program for granting us access to TPUv3 units and GCP credits, respectively.

\bibliography{anthology,custom}
\bibliographystyle{acl_natbib}
\newpage

\appendix

\section{Additional dataset information}
\label{sec:appendixA}
\makeatletter
\setlength{\@fptop}{0pt}
\makeatother
\vspace*{-\baselineskip}
\begin{table} [H]
\setlength{\tabcolsep}{1pt}
\centering
\small
\begin{tabular}{lcc}
\hline
\textbf{Category} & \textbf{Positive} & \textbf{Negative} \\
\hline
Intraventricular & 22 (\%1.1) & 1978 (\%98.9) \\
Gliosis & 54 (\%2.7) & 1946 (\%97.3) \\
Epidural & 51 (\%2.55) & 1949 (\%97.45) \\
Hydrocephalus & 70 (\%3.5) & 1930 (\%96.5) \\
Encephalomalacia & 177 (\%8.85) & 1823 (\%91.15) \\
Chronic ischemic changes & 951 (\%47.55) & 1049 (\%52.45) \\
Lacuna & 138 (\%6.9) & 1862 (\%93.1) \\
Leukoaraiosis & 49 (\%2.45) & 1951 (\%97.55) \\
Mega cisterna magna & 15 (\%0.75) & 1985 (\%99.25) \\
Meningioma & 39 (\%1.95) & 1961 (\%98.05) \\
Subarachnoid Bleeding & 209 (\%10.45) & 1791 (\%89.55) \\
Subdural & 227 (\%11.35) & 1773 (\%88.65) \\
No Findings & 299 (\%14.95) & 1701 (\%85.05) \\

\hline
\end{tabular}
\caption{Distribution of positive and negative frequencies for each label in the dataset}
\label{table:label_distribution}
\end{table}

\section{Pre-training and fine-tuning hyperparameters}
\label{sec:appendixB}

\vspace*{-\baselineskip}
\begin{table}[H]
\setlength{\tabcolsep}{1pt}
\centering
\begin{tabular}{lc}
\hline
\textbf{Hyperparameters} & \textbf{Values}\\
\hline
Learning rate & 1e-4 \\ 
Batch size & 256 \\ 
Optimizer & Adam \\ 
$\beta_1$ & 0.9 \\ 
$\beta_2$ & 0.999 \\ 
Warmp up steps & 10000 \\ 
Max sequence length & 512 \\
Max prediction per seq & 76\\
Masked MLM probability & 0.15\\
epoch & 1000000\\
\hline
\end{tabular}
\caption{Pre-training configuration for BERT models.}
\label{tab:pretraining}
\end{table}

\begin{table} [h]
\setlength{\tabcolsep}{1pt}
\centering
\begin{tabular}{lcc}
\hline
\textbf{Parameters} & \textbf{TurkRadBERT-sim}\\
\hline
Learning rate & 5e-5 \\ 
Batch size & 32 \\ 
Optimizer & Adam \\ 
Max sequence length & 512 \\
epoch & 20\\
\hline
\end{tabular}
\caption{Fine-tuning configuration for TurkRadBERT-sim family}
\label{tab:finetuningsim}
\end{table}

\begin{table} [H]
\centering

\begin{tabular}{lcc}
\hline
\textbf{Parameters} & \textbf{TurkRadBERT-task}\\
\hline
Learning rate & 3e-5 \\ 
Batch size & 32 \\ 
Optimizer & Adam \\ 
Max sequence length & 512 \\
epoch & 15\\
\hline
\end{tabular}
\caption{Fine-tuning configuration for TurkRadBERT-task family}
\label{tab:finetuningtask}
\end{table}

\end{document}